\documentclass[10pt,twocolumn,letterpaper]{article}

\usepackage[utf8x]{inputenc}

\usepackage{iccv}
\usepackage{times}
\usepackage{amsmath}
\usepackage{amssymb}
\usepackage{bbold}
\usepackage{latexsym}
\usepackage{multicol}
\usepackage{booktabs}
\usepackage{graphicx}
\usepackage{soul}
\usepackage{xcolor}
\usepackage{rotating}
\usepackage{booktabs}

\usepackage[breaklinks=true,bookmarks=false]{hyperref}

\iccvfinalcopy % *** Uncomment this line for the final submission

\def\iccvPaperID{35} % *** Enter the ICCV Paper ID here

\ificcvfinal\fi

\newcommand\floor[1]{\lfloor#1\rfloor}

\definecolor{DarkGreen}{rgb}{0.0, 0.2, 0.0}

\newcommand{\comment}[1]{}

\usepackage[normalem]{ulem} % for \sout

\begin{document}

%%%%%%%%% TITLE
\title{LU-Net: An Efficient Network for 3D LiDAR Point Cloud Semantic Segmentation Based on End-to-End-Learned 3D Features and U-Net}

% (vincent: ) I added " by Projection of 3D Features " to prevent people to think it is only U-Net on Lidar data

\author{Pierre Biasutti$^{1,2,3,4}$, Vincent Lepetit$^{1}$, Mathieu Brédif$^{3}$, Jean-François Aujol$^{2}$, Aurélie Bugeau$^{1}$\\
$^1$\textit{Univ. Bordeaux, CNRS,  Bordeaux INP, LaBRI, UMR 5800, F-33400, Talence, France}\\
$^2$\textit{Univ. Bordeaux, IMB, INP, CNRS, UMR 5251, F-33400 Talence, France}\\
$^3$\textit{Univ. Paris-Est, LASTIG GEOVIS, IGN, ENSG, F-94160 Saint-Mandé, France}\\
$^4$\textit{GEOSAT, Pessac, France}\\
pierre.biasutti@labri.fr}

\maketitle

\ificcvfinal\thispagestyle{empty}\fi

%%%%%%%%%%%%%%%%%%%%%%%%%%%%%%%%%%%%%%%%%%%%%%%%%%%%%%%%%%%%%%%%%%%%%%%%%%%%%%%%
\begin{abstract}
We propose LU-Net---for LiDAR U-Net, a new method for the semantic segmentation of a 3D LiDAR point cloud. Instead of applying some global 3D segmentation method such as PointNet, we propose an end-to-end architecture for LiDAR point cloud semantic segmentation that efficiently solves the problem as an image processing problem. We first extract high-level 3D features for each point given its 3D neighbors. Then, these features are projected into a 2D multichannel range-image by considering the topology of the sensor. Thanks to these learned features and this projection, we can finally perform the segmentation using a simple U-Net segmentation network, which performs very well while being very efficient. In this way, we can exploit both the 3D nature of the data and the specificity of the LiDAR sensor. This approach outperforms the state-of-the-art by a large margin on the KITTI dataset, as our experiments show. Moreover, this approach operates at 24fps on a single GPU. This is above the acquisition rate of common LiDAR sensors which makes it suitable for real-time applications.
\end{abstract}

% We propose LU-Net (for LiDAR U-Net), for the semantic segmentation of a 3D LiDAR point cloud. Instead of applying some global 3D segmentation method such as PointNet, we propose an end-to-end architecture for LiDAR point cloud semantic segmentation that efficiently solves the problem as an image processing problem. First, we use a high-level 3D feature extraction module to compute 3D local features for each point given its neighbors. Then, these features are projected into a 2D multichannel range-image by considering the topology of the sensor. This range-image later serves as the input to a U-Net segmentation network, which is a simple architecture yet enough for our purpose. In this way, we can exploit both the 3D nature of the data and the specificity of the LiDAR sensor. This approach efficiently bridges between 3D point cloud processing and image processing as it outperforms the state-of-the-art by a large margin on the KITTI dataset, as our experiments show. Moreover, this approach operates at 24fps on a single GPU. This is above the acquisition rate of common LiDAR sensors which makes it suitable for real-time applications.

%%%%%%%%%%%%%%%%%%%%%%%%%%%%%%%%%%%%%%%%%%%%%%%%%%%%%%%%%%%%%%%%%%%%%%%%%%%%%%%%
\section{Introduction}

The recent interest for autonomous systems has motivated many computer vision works over the past years. The importance of accurate perception models is a crucial step towards system automation, especially for mobile robots and autonomous driving. Modern systems are equipped with both optical cameras and 3D sensors, mostly LiDAR sensors. These sensors are now essential components of  perception systems as they enable direct space measurements, providing an accurate 3D representation of the scene. However, for most automation-related tasks, raw LiDAR point clouds require further processing in order to be used. In particular, point clouds with accurate semantic segmentation provide a higher level of representation of the scene that can be used in various applications such as obstacle avoiding, road inventory, or object manipulation.

\begin{figure}
    \centering
    \newcommand{\szw}{0.45}
    \begin{tabular}{r@{ }c}
        \begin{turn}{90}\tiny{\quad prediction}\end{turn} & \includegraphics[width=\szw\textwidth]{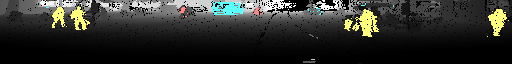} \\
        \begin{turn}{90}\tiny{\; groundtruth}\end{turn} & \includegraphics[width=\szw\textwidth]{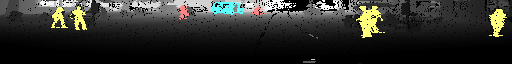} \\
        \begin{turn}{90}\tiny{\quad \quad \quad \quad \quad prediction}\end{turn} &\includegraphics[width=\szw\textwidth]{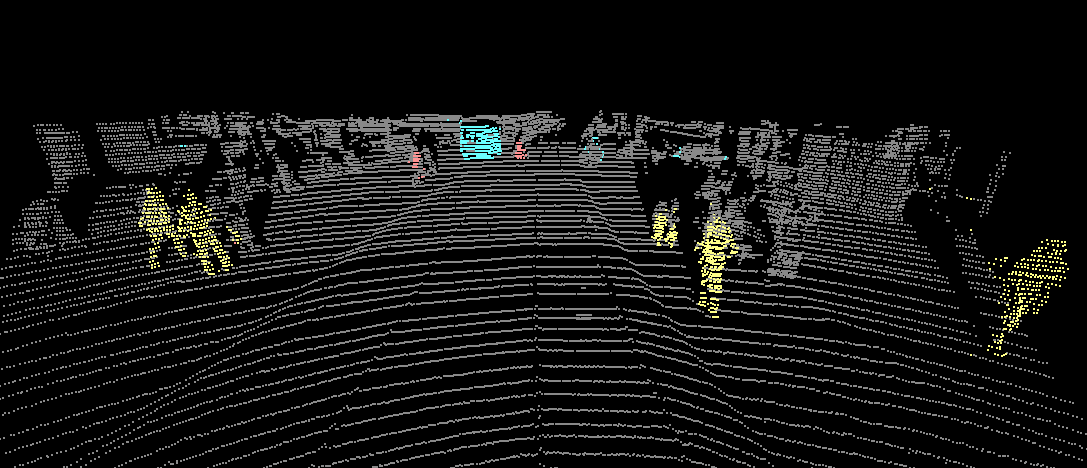} \\
        \begin{turn}{90}\tiny{\quad \quad \quad \quad \quad groundtruth}\end{turn} &\includegraphics[width=\szw\textwidth]{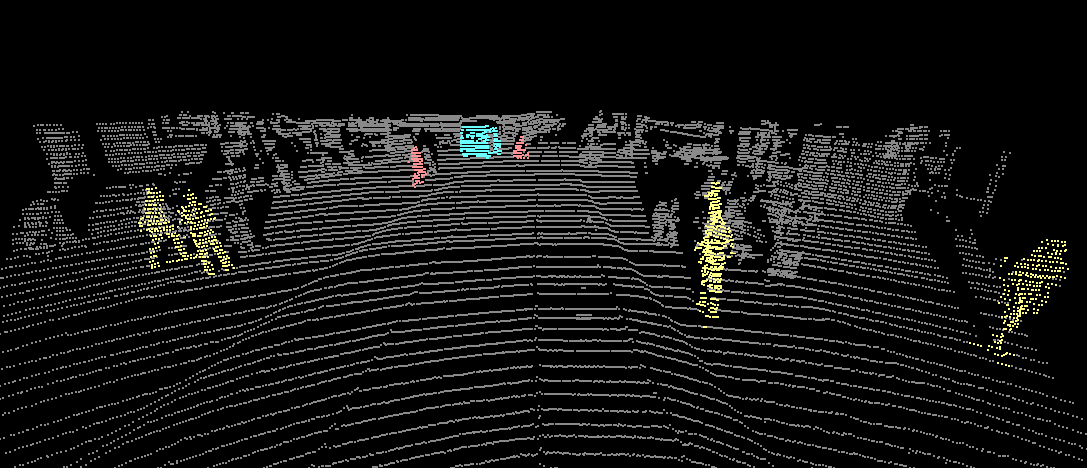} \\
    \end{tabular}
    \caption{The top two images show the segmentation of LiDAR data obtained with our method, and the groundtruth segmentation, seen in the sensor topology. The bottom two images show the same segmentations from a different point of view.}
    \label{fig:front}
\end{figure}

% Result of the range-image semantic segmentation produced by the proposed method. The first two results show the prediction of the proposed model and the groundtruth respectively, seen in the sensor topology. The last two results show the same prediction and groundtruth in 3D.

This paper focuses on the semantic segmentation of 3D LiDAR point clouds. Given a point cloud acquired with a LiDAR sensor, we aim at estimating a label for each point that belongs to objects of interest in urban environments (such as cars, pedestrians and cyclists). The traditional pipelines used to tackle this problem consider ground removal, clustering of remaining structures, and classification based on handcrafted features extracted on each clusters~\cite{himmelsbach2008lidar,feng2014fast}. The segmentation can be improved with variational models~\cite{landrieu2018large}. These methods are often hard to tune as handcrafted features usually require tuning many parameters, which is likely to be data dependent and therefore hard to use in a general scenario. Finally, although the use of regularization can lead to visual and qualitative improvements, it often leads to a large increase of the computational time.

Recently, deep-learning approaches have been proposed to overcome the difficulty of tuning handcrafted features. This has become possible with the arrival of large 3D annotated datasets such as the KITTI 3D object detection dataset~\cite{geiger2012are}. Many methods have been proposed to segment the point cloud by directly operating in 3D~\cite{qi2017pointnet} or on a voxel-based representation of the point cloud~\cite{zhou2018voxelnet}. However, this type of methods either needs very high computational power, or are not able to process the amount of points acquired in a single rotation of a sensor. Even more recently, faster approaches have been proposed~\cite{wu2018squeezeseg,wang2018pointseg}. They rely on a 2D representation of the point cloud, called range-image~\cite{biasutti2018range}, which can be used as the input of a convolutional neural network. Thus, the processing time as well as the required computational power can be kept low, as these range-images consist in low resolution, multichannel images. Unfortunately, the choice of input channels, as well as the difficulty of processing geo-spatial information using only 2D convolutions have limited the results of such approaches, which have not yet achieved good enough scores for practical use, especially on small objects classes such as cyclists or pedestrians.

%that are practically useful on certains classes.

In this paper, we propose LU-Net---for LiDAR U-Net---an end-to-end model for the semantic segmentation of 3D LiDAR point clouds. LU-Net benefits from a high-level 3D feature extraction module that can embed 3D local features in 2D range-images, which can later be efficiently used in a U-Net segmentation network. We demonstrate that, beside being a simple and efficient method, LU-Net largely outperforms state-of-the-art range-image methods, as shown in Figure~\ref{fig:front}.

The rest of the paper is organized as follows: We first discuss previous works on point cloud semantic segmentation, including methods designed for processing LiDAR data. We then detail our approach, and evaluate it on the KITTI dataset against state-of-the-art methods and discuss the results.

% (vincent: ) This does not provide much information:
% Finally a conclusion is drawn.

\section{Related Work}

In this section, we discuss previous works on image semantic segmentation as well as 3D point cloud semantic segmentation below.

\subsection{Semantic Segmentation for Images}

Semantic segmentation of images has been the subject of many works in the past years. Recently, deep learning methods have largely outperformed previous ones. The method presented in \cite{long2015fully} was the first to propose an accurate end-to-end network for semantic segmentation. This method is based on an encoder in which each scale is used to compute the final segmentation. Only a few month later, the U-Net architecture \cite{ronneberger2015u}  was proposed for the semantic segmentation of medical images. This method is an encoder-decoder able to provide highly precise segmentation. These two methods have largely influenced recent works such as DeeplabV3+~\cite{chen2018deeplabv3plus} that uses dilated convolutional layers and spatial pyramid pooling modules in an encoder-decoder structure to improve the quality of the prediction. Other approaches explore multi-scale architectures to produce and fuse segmentations performed at different scales \cite{lin2017refinenet,zhao2018icnet}. Most of these methods are able to produce very accurate results, on various types of images~(medical, outdoor, indoor).
The survey \cite{briot2018analysis} of CNNs methods for semantic segmentation provides a deep analysis of some recent techniques. This work demonstrates that a combination of various components would most likely improve segmentation results on wider classes of objects.

\subsection{Semantic Segmentation of Point Clouds}

\paragraph{3D-based methods.} As mentioned above, the first approaches for point cloud semantic segmentation were done using heavy pipelines, composed of many successive steps such as: ground removal, point cloud clustering, feature extraction as presented in \cite{himmelsbach2008lidar,feng2014fast}.
However, as mentioned above, these methods often require many parameters and they are therefore hard to tune. In \cite{landrieu2019point},  a deep-learning approach is used to extract features from the point cloud. Then, the segmentation is done using a variational regularization. Another approach presented in \cite{qi2017pointnet} proposes to directly input the raw 3D LiDAR point cloud to a network composed of a succession of fully-connected layers to classify or segment the point cloud. However, due to the heavy structure of this architecture, it is only suitable for small point clouds. Moreover, processing 3D data often increases the computational time due to the dimension of the data (number of points, number of voxels), and the absence of spatial correlation. To overcome these limitations, the methods presented in \cite{li20173d} and \cite{zhou2018voxelnet} propose to represent the point cloud as a voxel-grid which can be used as the input of a 3D CNN. These methods achieve satisfying results for 3D detection. However, semantic segmentation would require a voxel-grid of very high resolution, which would increase the computational cost as well as the memory usage.

\begin{figure*}[!t]
    \centering
    \includegraphics[width=0.95\textwidth]{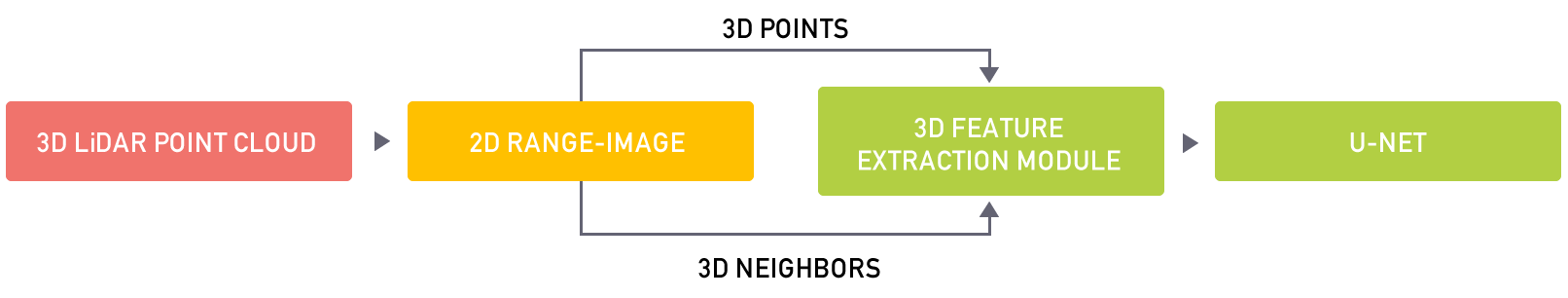}
    \caption{Proposed pipeline for 3D LiDAR point cloud semantic segmentation. First, the topology of the sensor is used to estimate the 8-connected neighborhood of each point. Then, each point and its neighbors are fed to the high-level 3D feature extraction module, which outputs a multichannel 2D range-image. The range-image is finally used as the input of a U-Net segmentation network.}
    \label{fig:pipeline}
\end{figure*}

\paragraph{Range-image based methods.} Recently, SqueezeSeg,  a novel approach for the semantic segmentation of a LiDAR point cloud represented as a spherical range-image~\cite{biasutti2018range}, was proposed. This representation allows to perform the segmentation by using simple 2D convolutions, which lowers the computational cost while keeping good accuracy. The architecture is derived from the SqueezeNet image segmentation method~\cite{iandola2016squeezenet}. The intermediate layers are "fire layers", \textit{i.e.} layers made of one squeeze module and one expansion module.
Later on, the same authors improved this method in \cite{wu2018squeezesegv2} by adding a context aggregation module and by considering focal loss and batch normalization to improve the quality of the segmentation.
A similar range-image approach was proposed in \cite{wang2018pointseg}, where a Atrous Spatial Pyramid Pooling~\cite{chen2018deeplab} and squeeze  reweighting  layer~\cite{hu2018squeeze} are added.
Finally, in \cite{biasutti2019riunet}, the authors offer to input a range-image directly to the U-Net architecture described in \cite{ronneberger2015u}. This method achieves results that are comparable to the state of the art of range-image methods with a much simpler and more intuitive architecture.
All these range-image methods succeed in real-time computation. However, their results often lack of accuracy which limits their usage in real scenarios.
\\

In the next section, we propose LU-Net: an end-to-end model for the accurate semantic segmentation of point clouds represented as range-images. We will show that it outperforms all other range-image methods by a large margin on the KITTI dataset, while offering a robust methodology for bridging between 3D LiDAR point cloud processing and 2D image processing.

\section{Methodology}
\label{sec:methodolgy}

In this section, we present our end-to-end model for the semantic segmentation of LiDAR point clouds inspired by the U-Net architecture~\cite{ronneberger2015u}. An overview of the proposed method is available in Figure~\ref{fig:pipeline}.

\subsection{Network input}

As mentioned above, processing raw LiDAR point clouds is computationally expensive. Indeed, these 3D point clouds are stored as unorganized lists of $(x,y,z)$ Cartesian coordinates. Therefore processing such data often involves preprocessing steps to bring spatial structure to the data. To that end, alternative representations, such as voxel grids or 2D pinhole projections in 2D images, are sometimes used, as discussed in the Related Work section. However, high resolution is often needed in order to represent enough details, which involves heavy memory costs.
Modern LiDAR sensors often acquire 3D points, following a strict sensor topology, from which we can build a dense 2D image~\cite{biasutti2018range}, the so-called range-image. The range-image offers a lightweight, structured and dense representation of the point cloud.

\subsection{Range-images}
\begin{figure}[b!]
	\centering
		\begin{tabular}{cc}
			\includegraphics[width=0.45\textwidth]{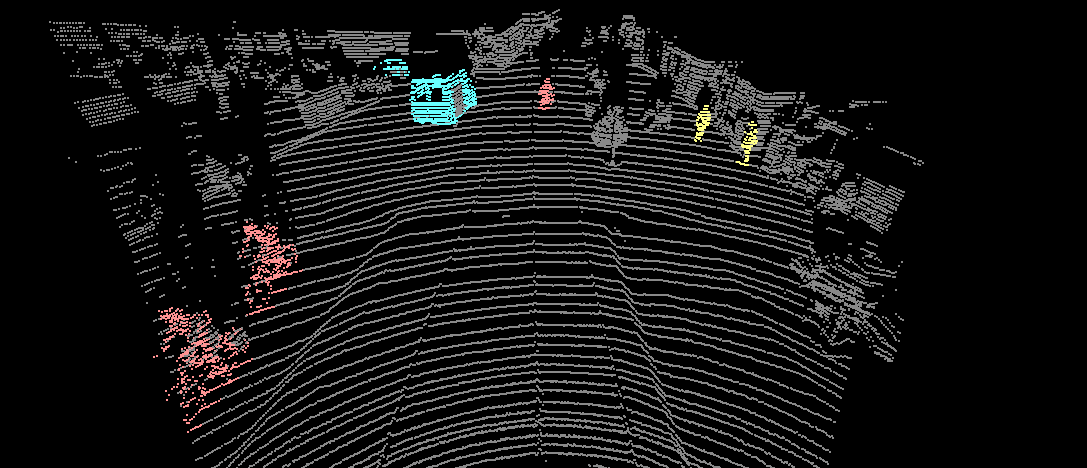}\\
			(a) \\
			\includegraphics[width=0.45\textwidth]{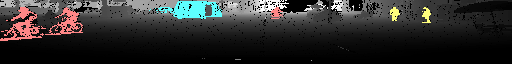}  \\
			(b)
 		\end{tabular}
		\caption{Turning a point cloud into a range-image. (a) A point cloud from the KITTI database~\cite{geiger2012are}, (b) the same point cloud as a a range-image. Note that the dark area in (b) corresponds to pulses with no returns. Colors correspond to groundtruth annotation, for better understanding.}
		\label{fig:range_map}
\end{figure}

Whenever the raw LiDAR data (with beam number) is not available, the point cloud has to be processed to extract the corresponding range-image.
As 3D LiDAR sensors acquire 3D points with a sampling pattern of a few number of scan lines and quasi uniform angular steps between samples, the acquisition follows a grid pattern that can be used to create a 2D image. Indeed, each point is defined by two angles and a depth, $(\theta, \phi, d)$ respectively, with steps of ($\Delta\theta, \Delta\phi$) between two consecutive positions. Each point $p_i$ of the LiDAR point cloud $P$ can be mapped to the coordinates $(x,y)$ with $x = \floor{\frac{\theta}{\Delta \theta}}, y = \floor{\frac{\phi}{\Delta \phi}}$ of a 2D range-image $u$ of resolution $H \times W = Card(P)$, where each channel represents a modality of the measured point.  A range-image is presented on Figure~\ref{fig:range_map}.

In perfect conditions, the resulting image is completely dense, without any missing data. However, due to the nature of the acquisition, some measurements are considered invalid by the sensor and they lead to empty pixels (no-data). This happens when the laser beam is highly deviated (\textit{e.g.} when going through a transparent material) or when it does not create any echo (\textit{e.g.} when the beam points in the sky direction). We propose to identify such pixels using a binary mask $m$ equal to $0$ for empty pixels and to $1$ otherwise. The analysis of multi-echo LiDAR scans is subject to future work.

\subsection{High-level 3D feature extraction module}

\begin{figure}[b!]
    \centering
    \includegraphics[width=0.4\textwidth]{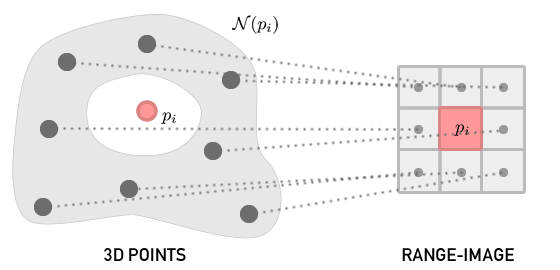}
    \caption{Illustration of the notation of the input of the feature extraction module. $p_i$ is the point, $\mathcal{N}(p_i)$ is the set of neighbors of $p_i$.}
    \label{fig:neighborhood}
\end{figure}

In \cite{wang2018pointseg}, \cite{wu2018squeezeseg} and \cite{wu2018squeezesegv2}, the authors use a 5-channel range-image as  input of their network. These 5 channels are made of the 3D coordinates ($x,y,z$), the reflectance ($r$) and the spherical depth ($d$). However, the analysis presented in \cite{biasutti2019riunet} showed that feeding a 2-channel range-image with only the reflectance and depth information to a U-Net architecture achieves comparable results to the state of the art.

In all these previous works, the choice of the number of channels of the range-image appears to be empirical. For each application, a complete study or a large set of experiments must be conducted to choose the best within all the possible combinations of channels. This is tedious and time consuming. To bypass such an expensive study, we propose in this paper a feature extraction module that is able to directly learn meaningful features adapted to the target application---here, semantic segmentation.

Moreover, processing geo-spatial information using 2D convolutional layers can cause issues in terms of data normalization as LiDAR sensors sampling typically decreases when acquiring farther points.

Inspired by the Local Point Embedder presented in \cite{landrieu2019point}, we propose a high-level 3D feature extraction module that is able to learn $N$ meaningful high-level 3D features for each point and to output a range-image with $N$ channels. Contrary to \cite{landrieu2019point}, our module exploits the range-image to directly estimate the neighbors of each points instead of using a pre-processing step. Moreover, our module outputs a range-image, instead of a point cloud, which can be used as input to a CNN.

\begin{figure}
    \centering
    \includegraphics[width=0.45\textwidth]{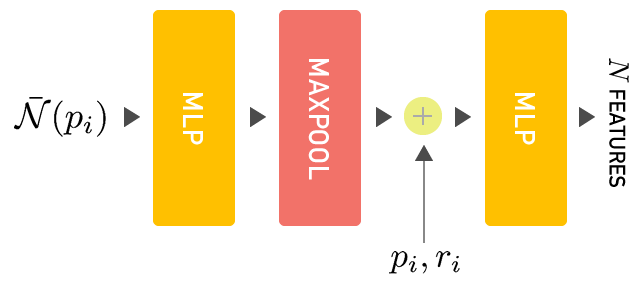}
    \caption{Architecture of the 3D feature extraction module. The output is an $1 \times N$ feature vector for each LiDAR point.}
    \label{fig:3d_feature_extractor}
\end{figure}

Given a point $p_i = (x,y,z)$, and $r_i$ its associated reflectance, we define $\mathcal{N}(p_i)$ the set of neighboring points of $p_i$ in the range-image (\textit{e.g.} the points that correspond to the 8-connected neighborhood of $p_i$ in the range-image). This set is illustrated Figure~\ref{fig:neighborhood}. We also define $\bar{\mathcal{N}}(p_i) = \{q - p_i \,|\, q \in \mathcal{N}(p_i)\}$ the set of neighbors in coordinates relative to $p_i$. Note that if either $p_i$ or $q$ is an empty pixel, then $q - p_i = (0, 0, 0)$.

Similarly to \cite{landrieu2019point}, the set of neighbors $\bar{\mathcal{N}}(p_i)$ is first processed by a multi-layer perceptron (MLP), which consists of a succession of linear, ReLU and batch normalization layers. The resulting set is then maxpooled to a point feature set, which is concatenated with $p_i$ and $r_i$. The resulting vector is processed through another MLP that outputs a vector of $N$ 3D features for each $p_i$. This module is illustrated in Figure \ref{fig:3d_feature_extractor}.

As linear layers can be done using $1\times1$ convolutional layers, the whole $P$ point cloud can be processed at once. In this case, the output of the 3D feature extraction module is a $\textrm{Card}(P) \times N$ matrix, which can then be reshaped to a $H \times W \times N$ range-image.

\subsection{Semantic segmentation}

\begin{figure}[h!]
    \centering
    \includegraphics[width=0.47\textwidth]{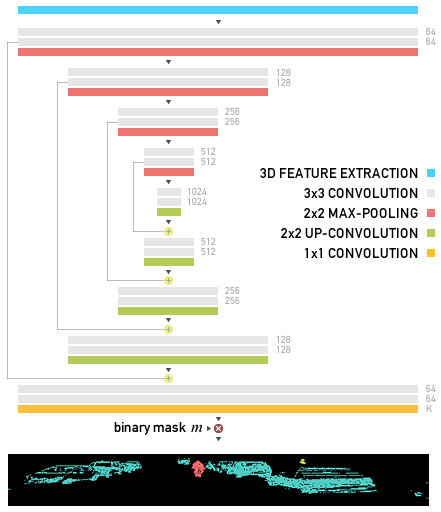}
    \caption{LU-Net architecture with the output of the 3D feature extraction module as the input (top) and the output segmented range-image (bottom).}
    \label{fig:unet_architecture}
\end{figure}

\paragraph{Architecture. } The U-Net architecture~\cite{ronneberger2015u} is an encoder-decoder. As illustrated in Figure~\ref{fig:unet_architecture}, the first half consists in the repeated application of two $3\times3$ convolutions followed by a rectified linear unit (ReLU) and a $2\times2$ max-pooling layer that downsamples the input by a factor 2. Each time a downsampling is done, the number of features is doubled to compensate for the loss of resolution. The second half of the network consists of upsampling blocks where the input is upsampled using $2\times2$ up-convolutions. Then, concatenation is done between the upsampled feature map and the corresponding feature map of the first half. This allows the network to capture global details while keeping fine details. After that, two $3\times3$ convolutions are applied followed by a ReLU. This block is repeated until the output of the network matches the dimension of the input. Finally, the last layer consists in a 1x1 convolution that outputs as many features as the wanted number of possible labels \textit{i.e.} $K$ 1-hot encoded.

\paragraph{Loss function. }
The loss function of our model is defined as a variation of the focal loss presented in \cite{lin2017focal}. Indeed, our model is trained on a dataset in which the number of example for each class is largely unbalanced. Using the focal loss approach helps improving the average score by few percents, as discussed later in Section \ref{sec:experiments}. First, we define the pixel-wise softmax for each label $k$:
\begin{align*}
p_k(x) = \frac{\textrm{exp}(a_k(x))}{\sum\limits_{k'=0}^{K}\textrm{exp}(a_{k'}(x))}
\end{align*}
where $a_k(x)$ is the activation for feature $k$ at the pixel position $x$. After that, we define $l(x)$ the groundtruth label of pixel $x$. We then compute the weighted focal loss as follows:
\begin{align*}
    E = \sum\limits_{x \in \Omega}-\mathbb{1}_{\{m(x) > 0\}}w(x)(1 - p_{l(x)}(x))^\gamma\textrm{log}(p_{l(x)}(x))
\end{align*}
where $\Omega$ is the domain of definition of $u$, $m(x)>0$ are the valid pixels, $\gamma = 2$ is the \textit{focusing} parameter and $w(x)$ is a weighting function introduced to give more importance to pixels that are close to a separation between two labels, as defined in \cite{ronneberger2015u}.

\paragraph{Training} We train the network with the Adam stochastic gradient optimizer and a learning rate set to $0.001$. We also use batch normalization with a momentum of 0.99 to ensure good convergence of the model.  Finally, the batch size is set to $4$ and the training is stopped after $10$ epochs.

\section{Experiments}
\label{sec:experiments}

We trained and evaluated LU-Net using the same experimental setup as the one presented in SqueezeSeg \cite{wu2018squeezeseg} as they provide range-images with segmentation labels exported from the 3D object detection challenge of the KITTI dataset \cite{geiger2012are}. They also provide the training / validation split that they used for their experiments, which contains $8057$ samples for training and $2791$ for validation and which can be used for a fair comparison between each result of each method.

We have manually tuned the number of layers $N$, \textit{i.e.} the number of 3D features learned for each points. On all our experiments, best semantic segmentation results were obtained by setting $N=3$. This  small amount of channels is enough to highlight the structure of the objects that are latter used in the U-Net in charge of the segmentation task. All results reported in this section are with this value.  Nevertheless, if using the high-level 3D feature extraction module for other applications, one should consider adapting this value.

\subsection{Comparison with the state of the art}

We compare the proposed method to 4 range-image based methods of the state of the art: PointSeg~\cite{wang2018pointseg}, SqueezeSeg~\cite{wu2018squeezeseg}, SqueezeSegV2~\cite{wu2018squeezesegv2}, and RIU-Net~\cite{biasutti2019riunet}. RIU-Net is a previous version of LU-Net we developed and was solely based on the raw reflectance and depth features instead of the 3D features learned in the end-to-end network of LU-Net. Similarly to \cite{wu2018squeezeseg} and \cite{wu2018squeezesegv2}, the comparison is done based on the Intersection-over-Union score:
\begin{align*}
    {\rm IoU}_l = \frac{|\rho_l \bigcap G_l|}{|\rho_l \bigcup G_l|}
\end{align*}
where $\rho_l$ and $G_l$ denote the predicted and groundtruth sets of points that belongs to label $l$ respectively.

\begin{table}[b!]
    \caption{Comparison (IoUs, $\%$) of LU-Net with the state of the art for the semantic segmentation of the KITTI dataset.}
    \begin{center}
    \begin{tabular}{cccccc}
        \toprule
            && \rotatebox{60}{Cars} & \rotatebox{60}{Pedestrians} & \rotatebox{60}{Cyclists} & \rotatebox{60}{Average} \\
        \toprule
        SqueezeSeg &\cite{wu2018squeezeseg}    & 64.6 & 21.8 & 25.1 & 37.2\\
        PointSeg &\cite{wang2018pointseg}      & 67.4 & 19.2 & 32.7 & 39.8\\
        RIU-Net &\cite{biasutti2019riunet} & 62.5 & 22.5 & 36.8 & 40.6 \\
        SqueezeSegv2 &\cite{wu2018squeezesegv2} & 73.2 & 27.8 & 33.6 & 44.9\\
        \midrule
        LU-Net   & & 72.7 & \textbf{46.9} & \textbf{46.5} & \textbf{55.4} \\
        \bottomrule
    \end{tabular}
    \end{center}
    \label{tab:iou}
\end{table}

The performance comparisons between LU-Net and state-of-the-art methods are displayed Table~\ref{tab:iou}. The first observation is that the proposed model outperforms existing methods in terms of average IoU by over $10$\%. In particular, the proposed model achieves better results on each of the classes compared to PointSeg, SqueezeSeg and RIU-Net. Our method also largely outperforms SqueezeSegV2 for both pedestrians and cyclists.

Our method is very similar to RIU-Net as both methods use a U-Net architecture with a range-image as input. While RIU-Net uses 2 channels---the reflectance and depth---LU-Net automatically extracts a N-dimensional high-level features per point thanks to the 3D feature extraction module. Table~\ref{tab:iou} demonstrates that using an additional network to automatically learn high-level features from the 3D point cloud largely improves the results, especially on classes that are less represented in the dataset.

\begin{figure}
    \centering
    \newcommand{\szw}{0.45\textwidth}
    \hspace{-0.1cm}\begin{tabular}{c}
        \includegraphics[width=\szw]{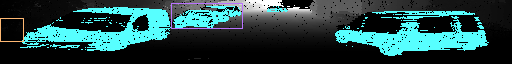} \\
        Ground truth\vspace{0.2cm}\\
        \includegraphics[width=\szw]{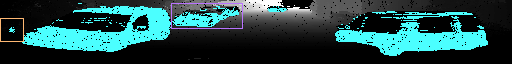} \\
        SqueezeSegV2 \protect\cite{wu2018squeezesegv2}\vspace{0.2cm}\\
        \includegraphics[width=\szw]{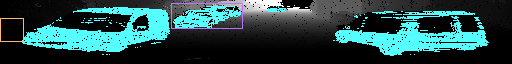} \\
        LU-Net \vspace{0.3cm}\\
        \includegraphics[width=\szw]{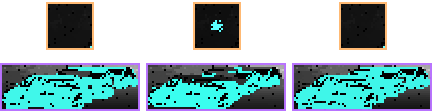} \\
        Zooms in the following order \\{\it Groundtruth, SqueezeSegV2, LU-Net}\vspace{0.1cm}

    \end{tabular}
    \caption{Visual comparison of the proposed model against SqueezeSegV2 \protect\cite{wu2018squeezesegv2} and the ground truth. Results are shown on the range-image where depth values are encoded with a grayscale map. Both SqueezeSegV2 and LU-Net globally achieve very satisfying results. Nervertheless, LU-Net is less subject to false positives than SqueezeSegV2, as can be seen in the orange areas and corresponding zooms. It also better segments farther objects such as the cars on the back of the scene in the purple rectangle, which reduces the amount of false-negatives, which are crucial for autonomous driving applications.}
    \label{fig:comparison}
\end{figure}

Figure~\ref{fig:comparison} presents visual results for SqueezeSegV2 and LU-Net. We here observe that visually, the results for cars are comparable. Nevertheless, by looking closer at the results, we observe that SqueezeSegV2 is more subject to false positives~(Figure~\ref{fig:comparison}, orange rectangle). Moreover, our method provides a better segmentation of the cars in the back of the scene,  compared to SqueezeSegV2~(Figure \ref{fig:comparison}, purple rectangle).

\subsection{Ablation study}

Table~\ref{tab:iou2} presents intermediate scores in order to highlight the contribution of some model components.

First, we analyse the influence of relative coordinates $\bar{\mathcal{N}}$ as input to the 3D feature extraction module (Figure~\ref{fig:3d_feature_extractor}).  We trained and tested the model using absolute coordinates $\mathcal{N}$. We name this version LU-Net w/o relative. As Table~\ref{tab:iou2}  shows, relative candidates provide better results than neighbors in absolute coordinates. We believe that by reading relative coordinates as input, the network learns high-level features characterizing the local 3D geometry of the point cloud, independently of its absolute position in the 3D environment. These absolute positions are re-introduced once this geometry is learned, {\it i.e.} before the second multi-layer perceptron of the 3D feature extraction module.

\begin{figure}
    \centering
    \includegraphics[width=0.48\textwidth]{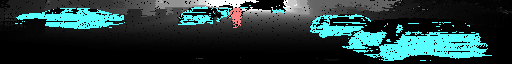} \\
    Groundtruth\vspace{0.2cm}\\
    \includegraphics[width=0.48\textwidth]{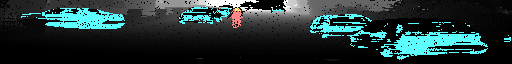} \\
    LU-Net w/o relative\vspace{0.2cm}\\
    \includegraphics[width=0.48\textwidth]{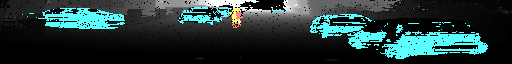} \\
    LU-Net w/o FL\vspace{0.2cm}\\
    \includegraphics[width=0.48\textwidth]{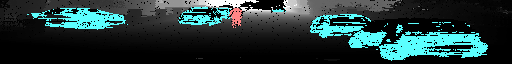} \\
    LU-Net\\
    \caption{Visual results of the ablation study. The use of neighbors in absolute coordinates results in incomplete segmentations of the objects compared to neighbors in relative coordinates. Moreover, the use of the focal loss~(FL) helps the network to better distinguish classes that have similar aspects, here, cyclists and pedestrians.}
    \label{fig:my_label}
\end{figure}

\begin{table}
    \caption{Ablation study for the semantic segmentation of the KITTI dataset. Results in terms of (IoUs, $\%$) for LU-Net w/o relative: which uses absolute coordinates $\mathcal{N}$ instead of relative $\bar{\mathcal{N}}$ as input to the feature extraction module;  LU-Net w/o FL : proposed model without focal-loss; LU-Net: proposed model with relative coordinates and focal-loss.
   }
    \begin{center}
    \begin{tabular}{lcccc}
        \toprule
            & \rotatebox{60}{Cars} & \rotatebox{60}{Pedestrians} & \rotatebox{60}{Cyclists} & \rotatebox{60}{Average} \\
        \toprule
        LU-Net w/o relative &  62.8 & 39.6 & 37.5 & 46.6\\
        LU-Net w/o FL         & \textbf{73.8} & 42.7 & 32.9 & 49.8 \\
        LU-Net    & 72.7 & \textbf{46.9} & \textbf{46.5} & \textbf{55.4} \\
        \bottomrule
    \end{tabular}
    \end{center}
    %LU-Net (global): uses $\cal{N}$ as neighbors, for comparison \\
    %LU-Net: proposed model without %focal-loss\\
    %LU-Net + FL: proposed model
    \label{tab:iou2}
\end{table}

For fair comparison, we also experimented using absolute coordinates $\mathcal{N}$ and adding a supplementary convolutional layer as the first layer. Indeed, we could expect this additional layer to characterize the  transformation from  absolute and local coordinates. Nevertheless, this architecture brought numerical instability while not managing to learn such transform, as it ended up with an average IoU of $30.6$\%.

\begin{figure*}
    \centering
    \newcommand{\szw}{0.65\textwidth}
    \hspace{-0.1cm}\begin{tabular}{c}
        \includegraphics[width=\szw]{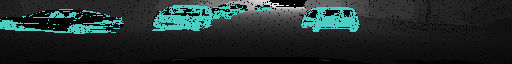}\\
        \includegraphics[width=\szw]{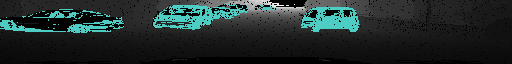} \\
        (a)\\

        \includegraphics[width=\szw]{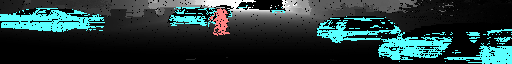}
        \\
        \includegraphics[width=\szw]{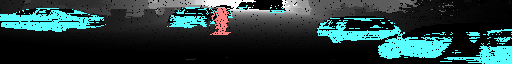}
        \\
        (b)\\

        \includegraphics[width=\szw]{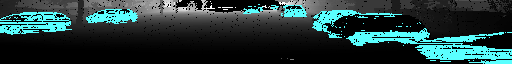} \\
        \includegraphics[width=\szw]{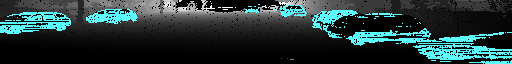} \\
        (c)\\

        \includegraphics[width=\szw]{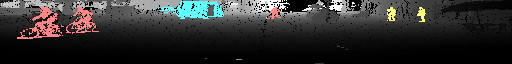}\\
        \includegraphics[width=\szw]{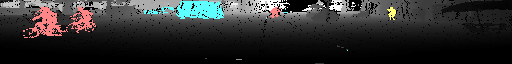}\\
        (d)\\

        \includegraphics[width=\szw]{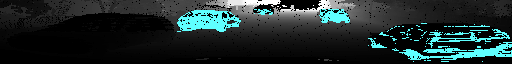}\\
        \includegraphics[width=\szw]{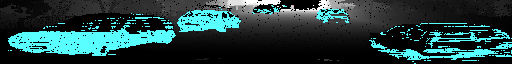}\\
        (e)
    \end{tabular}
    \caption{Results of the semantic segmentation of the proposed method (bottom) and groundtruth (top). Results are shown on the range-image where depth values are encoded with a grayscale map. Labels are associated to colors as follows: blue for  cars, red for  cyclists and lime for  pedestrians.}
    \label{fig:results}
\end{figure*}

Next, we analyse the influence of the focal-loss. As seen in Table~\ref{tab:iou2}, the use of focal-loss largely improves the scores on both cyclists and pedestrians. This is related to the imbalance between each class in the dataset, where there are $10$ times more car examples than cyclists or pedestrians.

\subsection{Additional results}
Apart from being convincing in terms of IoUs, the results produced by our method are also very convincing visually, as it is demonstrated Figure~\ref{fig:front} and \ref{fig:results}. Our segmentations are very close to those of the groundtruth. In Figure~\ref{fig:results}d), one of the pedestrians was not detected. When looking closely at the depth values in the range-image, this pedestrian is in fact hardly visible. It is also the case in the reflectance image. This is also related to the resolution of the sensor as only few points fall on the pedestrian, and could probably be solved by adding an external modality such as an optical image.

In Figure~\ref{fig:results}e), a car in the foreground is missing from the groundtruth, this causes the IoU to drop from $89.7$\% when ignoring this region of the image, down to $36.4$\%. Thus, removing examples with wrong or missing annotations in the dataset could lead to better results on LU-Net as well as on other methods. However, due to the amount of examples in the dataset, having a perfect annotation is practically very difficult.\\

Finally, LU-Net is able to operate at 24 frames per second on a single GPU. This is a lower frequency compared to other systems, yet still above the frame rate of the LiDAR sensor~(10fps for the Velodyne HDL-64e). Moreover, our system uses only a few more parameters than RIU-Net for a significant improvement in terms of IoU scores.

\section{Conclusion}
In this paper, we have presented LU-Net, an end-to-end model for the semantic segmentation of 3D LiDAR point clouds. Our method efficiently creates a multi-channel range-image using a learned 3D feature module. This range-image later serves as the input of a U-Net architecture. We show that this methodology efficiently bridges between 3D point cloud processing and image processing.
The resulting method is simple, but yet provides very high quality results far beyond existing state-of-the-art methods.

The current method relies on the focal loss function. We plan to study possible spatial regularization schemes within this loss function. Finally, fusion of LiDAR and optical data would probably enable reaching a higher level of accuracy.

\section{Acknowledgement}
The authors thank GEOSAT for funding part of this work. This project has also received funding from the European Union’s Horizon 2020 research and innovation programme under the Marie Skłodowska-Curie grant agreement No 777826.

\bibliographystyle{plain}

\end{document}